\documentclass{article}

\usepackage{PRIMEarxiv}

\usepackage[utf8]{inputenc} % allow utf-8 input
\usepackage[T1]{fontenc}    % use 8-bit T1 fonts
\usepackage{hyperref}       % hyperlinks
\usepackage{url}            % simple URL typesetting
\usepackage{booktabs}       % professional-quality tables
\usepackage{amsfonts}       % blackboard math symbols
\usepackage{nicefrac}       % compact symbols for 1/2, etc.
\usepackage{microtype}      % microtypography
\usepackage{lipsum}
\usepackage{fancyhdr}       % header
\usepackage{graphicx}       % graphics
\graphicspath{{media/}}     % organize your images and other figures under media/ folder
\usepackage{amsmath}
\usepackage{amsthm}
\newtheorem{theorem}{Theorem}
\newtheorem{lemma}{Lemma}
\newtheorem{corollary}{Corollary}

\usepackage{graphicx}
\usepackage{float}
\usepackage{algorithm}
\usepackage{algorithmicx}
\usepackage{algpseudocode}
\usepackage{booktabs}
\usepackage{multirow}
\usepackage{makecell}
\usepackage{xcolor}
\title{CAWR: Corruption-Averse Advantage-Weighted Regression for Robust Policy Optimization
}

\author{
  Ranting Hu \\
  ISTBI \\
  Fudan University \\
  Shanghai\\
  \texttt{rthu22@m.fudan.edu.cn} \\
  \texttt{amiahrt@163.com}\\
}

\begin{document}
\maketitle

\begin{abstract}
Offline reinforcement learning (offline RL) algorithms often require additional constraints or penalty terms to address distribution shift issues, such as adding implicit or explicit policy constraints during policy optimization to reduce the estimation bias of functions. This paper focuses on a limitation of the Advantage-Weighted Regression family (AWRs), i.e., the potential for learning over-conservative policies due to data corruption, specifically the poor explorations in suboptimal offline data. We study it from two perspectives: (1) how poor explorations impact the theoretically optimal policy based on KL divergence, and (2) how such poor explorations affect the approximation of the theoretically optimal policy. We prove that such over-conservatism is mainly caused by the sensitivity of the loss function for policy optimization to poor explorations, and the proportion of poor explorations in offline datasets. To address this concern, we propose Corruption-Averse Advantage-Weighted Regression (CAWR), which incorporates a set of robust loss functions during policy optimization and an advantage-based prioritized experience replay method to filter out poor explorations. Numerical experiments on the D4RL benchmark show that our method can learn superior policies from suboptimal offline data, significantly enhancing the performance of policy optimization.
\end{abstract}

% keywords can be removed
% \keywords{Offline reinforcement learning \and Over-conservatism \and Robust loss}

\section{Introduction}

Standard Reinforcement Learning (RL) can solve complex optimization problems, but faces practical challenges in many real-world scenarios: (1) high-cost exploration in real-world scenarios \cite{review, dialog}; (2) exponentially growing policy spaces in multistep planning \cite{R2D3}; and (3) sparse reward estimation difficulties \cite{HER}.
In fact, offline RL addresses these by learning from static datasets, eliminating exploration risks, and enabling flexible data usage. Through policy constraints and resampling, it handles large action spaces and sparse rewards effectively. 

Due to the "off-line" nature of offline reinforcement learning, classical offline RL algorithms require introducing additional constraints or penalty terms into standard RL algorithms \cite{DQN, Double_Q_learning, DDPG, TD3, PG, AC, A3C, PPO} to address the distributional shift problem. Typical approaches include policy constraints \cite{TD3_BC, BPPO, RWR, AWR, AWAC} and value regularization \cite{CQL, MCB, PessORL}.
Generally, these improvement methods mainly incorporate explicit or implicit policy constraints during policy optimization. They constrain the learned policy to remain close to the behavior policy (i.e., the policy used to collect the offline dataset), thereby preventing the agent from making overly "optimistic" estimations about out-of-distribution regions.

However, data corruption significantly impacts the performance of offline RL algorithms\cite{AW, DW, CQL_ReDS}, and excessive constraints may lead to overly conservative policies that fail to fully utilize good explorations (i.e., high-advantage actions) within the dataset - a phenomenon we term as \textbf{over-conservatism problem}.

For example, value regularization approaches, such as Conservative Q-Learning (CQL) \cite{CQL}, employ regularization terms to enforce implicit policy constraints. These terms aim to maximize Q values for in-distribution actions while minimizing values for out-of-distribution actions, biasing the agent toward in-distribution choices. However, this uniform treatment of all in-distribution actions disregards their varying qualities, consequently still leading to over-conservatism. Indeed, Singh et al. \cite{CQL_ReDS} demonstrated that when the variability in the dataset behavior across different states is large, distribution constraint algorithms could only attain a limited policy improvement. Their proposed CQL(ReDS) algorithm addresses this by incorporating penalties for poor explorations (i.e., low-advantage actions) in the value regularization term, directing the agent's focus toward good explorations under the behavior policy. Yeom et al. \cite{EPQ} observed that the CQL regularization term can introduces unnecessary bias, since Q function estimates are inherently more accurate for densely distributed samples. Their Exclusively Penalized Q-learning (EPQ) algorithm adapts regularization strength across samples, reducing constraints on high-density regions to improve model generalization.

Similarly, Twin Delayed Deep Deterministic Policy Gradient plus Behavior Cloning (TD3+BC) \cite{TD3_BC} incorporates a behavior cloning term \cite{BC} during policy optimization, indiscriminately forcing the learned policy to mimic the actions of the behavior policy. When the dataset is suboptimal, this constraint can cause the agent to overlook good explorations, resulting in overly conservative policies. To addresses this, Adaptive Advantage-Guided Policy Regularization (A2PR) \cite{A2PR} replaced the behavior policy with a superior policy estimated via Conditional Variational Auto-Encoder (CVAE) \cite{CVAE}, which preferentially selects good explorations and leading the agent to imitate these actions. Alternative approaches like Latent-variable Advantage-weighted Policy Optimization (LAPO) \cite{LAPO} and Advantage-Aware Policy Optimization (A2PO) \cite{A2PO} directly employ CVAE as the policy function, and A2PO further introducing advantage-based conditional inputs — analogous to return-based supervised learning, enabling the agent to distinguish between good and poor explorations, thus preventing over-conservatism.

Evidently, these algorithms primarily mitigate the over-conservatism problem by reducing the influence of poor explorations on policy optimization. However, existing research still lacks studies on the over-conservatism problem in the Advantage-Weighted Regression family (AWRs) \cite{RWR, AWR, AWAC, IQL, CRR}, which employs the Kullback–Leibler (KL) divergence between policy distributions as constraints, essentially approximating the theoretically optimal policy through advantage weighting. Notably, Implicit Q-Learning (IQL) \cite{IQL} (one of the AWRs) currently stands as one of the state-of-the-art algorithms in offline reinforcement learning. Although Yang et al. \cite{Robust_IQL} emploied the Huber loss to handle data corruption, this work only limited to esuring robust estimation for policy evaluation functions.

Indeed, AWRs also applies uniform constraint strength to all samples during policy optimization, which may lead to overly conservative learned policies. We aim to investigate this issue from two perspectives:

\begin{enumerate}
    \item \textit{How do poor explorations impact the theoretically optimal policy based on KL divergence constraints?}
    \item \textit{How do poor explorations affect the approximation of the theoretically optimal policy?}
\end{enumerate}

Through theoretical analysis of these two issues, we identify two primary factors contributing to the over-conservatism problem in AWRs: (1) the sensitivity of the approximation loss function to poor explorations, and (2) the proportion of poor explorations in offline data. 
Therefore, we propose the \textbf{Corruption-Averse Advantage-Weighted Regression (CAWR)} algorithm, which addresses these issues by implementing a set of robust loss functions to mitigate the impact of poor explorations on approximation, and uses an advantage-based prioritized experience replay method to reduce poor exploration in training data. 

We provide theoretical validation for our approach and experimentally evaluate the algorithm on the D4RL benchmark. The results show that employing more robust loss functions and reducing poor explorations in training data both enhance policy optimization performance, and RWAR outperforms IQL across multiple D4RL benchmark datasets, achieving superior policy optimization performance.

\section{Preliminaries}
\subsection{Markov decision process and offline RL}

In reinforcement learning, each trial is considered an episode where at the beginning of each episode, the agent is reset to an initial state $s_0$ sampled from the probability distribution $d_0(s)$, and at each subsequent step, the agent selects an action $a$ based on the current state $s$, receiving feedback from the environment including the reward $r$ and next state $s'$ until the episode terminates. Each collected tuple $(s, a, r, s')$ is called a transition, and the set of all transitions from an episode $\{(s_t, a_t, r_t, s_{t+1})\}_{t=0}^T$ forms a trajectory. Furthermore, reinforcement learning formalizes this interaction process as a Markov Decision Process (MDP) denoted by $(\mathcal{S}, \mathcal{A}, p, r, \gamma)$, where $\mathcal{S}$ represents the state space, $\mathcal{A}$ the action space, $p$ the state transition probability function defined as $p(s'|s, a) = P(s_{t+1}=s'|s_t=s, a_t=a)$, $r$ the bounded reward function satisfying $r(s, a) \leq R_{\max}$ for all $s \in \mathcal{S}$ and $a \in \mathcal{A}$, and $\gamma \in (0,1]$ the discount factor used to compute discounted returns, with higher values of $\gamma$ indicating greater emphasis on long-term rewards in the return calculation.

The objective of reinforcement learning is to enable the agent to learn an optimal policy that maximizes the return over episodes, formally expressed as:
\begin{equation}\label{eq1} 
\max_{\pi}\ \mathbb{E}_\pi \left[\sum_{t=0}^\infty \gamma^t r(s_t, a_t)\right], 
\end{equation}
where $\pi$ represents the policy function that characterizes the probability of the agent selecting action $a$ in state $s$, i.e., $\pi(a\vert s)=P(\text{action}=a\vert \text{state}=s)$, with the expectation $\mathbb{E}_\pi$ taken over trajectories generated by $\pi$ through state-action space.

For offline RL, the training is operated on a static dataset $D={(s_i, a_i, r_i, s'_i)}_{i=1}^N$ collected through interactions between the environment and a behavior policy $\pi_\beta$ (which may represent an ensemble of multiple policies), where the agent's objective is to estimate an optimal policy $\pi$ solely from this pre-collected dataset. 

\subsection{Advantage-Weighted Regression}

This paper collectively refers to RWR (Reward-Weighted Regression) \cite{RWR}, AWR (Advantage-Weighted Regression) \cite{AWR}, AWAC (Advantage-Weighted Actor Critic) \cite{AWAC}, IQL (Implicit Q-Learning) \cite{IQL}, etc., as the family of Advantage-Weighted Regression algorithms, since these methods share similar policy optimization formulations. Taking AWAC as an example, this algorithm employs the KL divergence as its policy constraint and solves the following optimization problem:
\begin{align}\label{eq2}
    \max_\pi\ &\mathbb{E}_{s\sim D, a\sim \pi(\cdot\vert s)}[A^{\pi_\beta}(s, a)]\\
    \text{s.t.}\ &\mathbb{E}_{s\sim D} D_{KL}(\pi\vert \vert \pi_\beta)(s)\le \epsilon.\nonumber
\end{align}
It can be observed that this formulation is mathematically equivalent to the optimization problem in Trust Region Policy Optimization (TRPO) algorithm \cite{TRPO}. However, instead of using the Natural Policy Gradient (NPG) method \cite{NPG} for solution, AWAC first derives the analytical solution to optimization problem (\ref{eq2}) through the Lagrange multiplier method \cite{Largrange}, i.e.,
\begin{lemma}(The theoretically optimal policy of AWRs)\label{lemma1}
    \begin{equation}\label{eq3}
        \pi_\beta^*(a\vert s) = \frac{1}{Z(s)}\pi_\beta(a\vert s)\exp{\left[\frac{1}{\lambda} A^{\pi_\beta}(s,a)\right]}.
    \end{equation}
    where $\lambda$ denotes the Lagrange multiplier and $Z(s)=\sum_{a\in \mathcal{A}(s)}\left[ \pi_\beta(a\vert s) \exp{\left(\frac{1}{\lambda} A^{\pi_\beta}(s,a)\right)}\right]$ serves as the normalization term that ensures the policy remains a valid probability distribution.
\end{lemma}
Equation (\ref{eq3}) reveals that as $\lambda \to +\infty$, the exponential term $\exp{\left[\frac{1}{\lambda} A^{\pi_\beta}(s,a)\right]} \to 1$, causing the optimal policy $\pi_\beta^*$ to degenerate to the behavior policy $\pi_\beta$. This indicates that stronger constraint intensity (larger $\lambda$) forces the learned policy to stay closer to $\pi_\beta$. Conversely, when $\lambda \to 0$, the policy converges to $\arg\max_a A^{\pi_\beta}(s,a)$, which corresponds to the solution of the unconstrained optimization problem.

It should be noted that while $\pi_\beta^*$ can be theoretically derived in this form, its practical implementation still requires numerical approximation. Therefore, AWAC estimates the optimal policy by minimizing the KL divergence between the parametric policy $\pi$ and the theoretically optimal $\pi_\beta^*$, formally solving:
\begin{equation}\label{eq4}
    \min_\pi\ \mathbb{E}_{s\sim D}[D_{KL} (\pi_\beta^* \vert \vert \pi)]. 
\end{equation}
By further deriving the above expression into a computable form using dataset $D$, we obtain AWAC's final policy optimization objective function as follows:
\begin{equation}\label{eq5}
    \min_\pi\ \mathbb{E}_{s,a\sim D}\left[-\log \pi(a\vert s)\exp{\left[\frac{1}{\lambda} A^{\pi_\beta}(s,a)\right]}\right].
\end{equation}
It can be observed that AWAC essentially functions as a behavior cloning (BC) \cite{BC} weighted by $\exp{\left[\frac{1}{\lambda} A^{\pi_\beta}(s,a)\right]}$. This is because when $A^{\pi_\beta}(s,a)>0$, the action $a$ is superior to $\pi_\beta$ and should be reinforced, resulting in a weight greater than 1; conversely, when $A^{\pi_\beta}(s,a)<0$, the action $a$ is inferior to $\pi_\beta$ and should be suppressed, yielding a weight less than 1. Notably, as $\lambda \to +\infty$, this formulation further degenerates into the cross-entropy between policy $\pi$ and behavior policy $\pi_\beta$, equivalent to the standard BC algorithm.

\section{How AWRs fails with poor explorations}
In general, an offline dataset $D$ typically contains both good and poor explorations, or trajectories collected by policies of varying quality. Specifically, the goodness of an exploration is determined by its advantage value $A^{\pi_\beta}(s,a)$. For a given advantage threshold $\epsilon>0$, if $A^{\pi_\beta}(s,a)>\epsilon$, the exploration $a$ is considered good exploration as it demonstrates $\epsilon$-level superiority over the average performance; otherwise, it is deemed poor exploration.

As AWRs can be seen as two parts - the theoretically optimal policy $\pi_\beta^*$ and the approximation of this optimal policy, we can analyze the over-conservatism issue in AWRs from two perspectives: (1) the impact of poor explorations on the theoretically optimal policy $\pi_\beta^*$, and (2) their influence on the approximation of this optimal policy.

\subsection{How poor explorations impact the theoretically optimal policy}

To analyze this impact, we first consider a scenario without policy constraints, where the KL divergence-based policy constraint is replaced with the entropy of the policy distribution:
\begin{align}\label{eq6}
    \max_\pi\ &\mathbb{E}_{s\sim D, a\sim \pi(\cdot\vert s)}[A^{\pi_\beta}(s, a)] + \lambda\cdot \mathcal{H}(\pi).
\end{align}
And then we obtain its solution, which we denote as the unbiased optimal policy:
\begin{lemma}[The unbiased optimal policy]\label{lemma0}
\begin{equation*}
    \pi^*(a\vert s) = \frac{1}{Z(s)}\exp{\left[\frac{1}{\lambda} A^{\pi_\beta}(s,a)\right]},
\end{equation*}
where $Z(s)=\sum_{a\in\mathcal{A}}\exp{\left[\frac{1}{\lambda}A^{\pi_\beta}(s,a)\right]}$ serves as the normalization term that ensures the policy remains a valid probability distribution. 
\end{lemma}
By comparing it with the constrained optimal policy $\pi_\beta^*$, we obtain the following theorem:
\begin{theorem}[A lower bound of the KL divergence between $\pi_\beta^*$ and $\pi^*$]\label{thm1}
The KL divergence between policy $\pi_\beta^*$ and the unbiased optimal policy $\pi^*$ satisfies the following inequality:
\begin{equation*}
    D_{KL}(\pi^*\vert \vert\pi_\beta^*) \ge \mathcal{H}(\pi^*, \pi_\beta)-\mathcal{H}(\pi_\beta, \pi^*).
\end{equation*}
\end{theorem}
The above theorem demonstrates that when the offline dataset $D$ contains significantly more poor explorations than good ones, the iteratively learned policy $\pi_\beta^*$ would become overly conservative and deviate substantially from the unbiased optimal policy $\pi^*$, resulting in the over-conservatism problem.

For instance, consider an action space with two actions $a_1$ and $a_2$, where $a_1$ is superior to $a_2$. Assume $\pi(a_1) > \pi(a_2) > \epsilon > 0$ holds. If the behavior policy $\pi_\beta$ predominantly generates poor explorations such that $\pi_\beta(a_1) \to 0$ and $\pi_\beta(a_2) \to 1$, then
\begin{align*}
    \mathcal{H}(\pi^*, \pi_\beta) &= -\pi^*(a_1)\log \pi_\beta(a_1)-\pi^*(a_2)\log \pi_\beta(a_2)\to +\infty\\
    \mathcal{H}(\pi_\beta, \pi^*)&=-\pi_\beta(a_1)\log \pi^*(a_1)-\pi_\beta(a_2)\log \pi^*(a_2)\to -\log \pi^*(a_2)<-\log\epsilon.
\end{align*}
Thus $D_{KL}(\pi^*\vert \vert\pi_\beta^*) \to +\infty$, i.e. $\pi_\beta^*$ fails to reach the unbiased optimal policy $\pi^*$.

\subsection{How poor explorations affect the approximation of the theoretically optimal policy}
To analyze the robustness of the loss function, we first rewrite the policy loss function (\ref{eq5}) as a simpler one, i.e.
\begin{equation}\label{eq7}
    \min_\mu\ \mathbb{E}_{s,a\sim D}\left[w_{s,a}F(\mu(s),a)\right],
\end{equation}
where $w_{s,a}=\exp{\left[\frac{1}{\lambda} A^{\pi_\beta}(s,a)\right]}$, $F(\mu(s),a)=-\log \pi(a\vert s)$, and $\mu(s)$ is the policy function that generate actions that satisfies $\mathbb{E}_{a\sim\pi(\cdot\vert s)} \left[a\right]=\mu(s)$.

We further define $\pi^+(a|s)$ as the policy corresponding to good exploration in state $s$, and $\pi^-(a|s)$ as the policy corresponding to poor exploration in state $s$, and suppose there exists $\epsilon\in[0,1]$ such that the behavior policy $\pi_\beta$ satisfies:
$$\pi_\beta=(1-\epsilon)\pi^+ +\epsilon\pi^-.$$
Also, we denote $\mu^*$ as the global optimal solution of Equation (\ref{eq7}), and $\mu^+$ the global optimal solution of Equation (\ref{eq7}) when the behavior policy is solely $\pi^+(a|s)$. Then we have the following theorem:

\begin{theorem}[An upper bound of the approximation bias in AWRs introduced by poor explorations]\label{thm2}
Suppose function $F(\cdot)$ is convex and twice differentiable at $\mu^+$, then for state $s$, the absolute difference between $\mu^*$ and $\mu^+$ obeys the following inequality:
\begin{equation*}
\vert \mu^*(s)-\mu^+(s)\vert \le \left( \mathbb{E}_{a\sim D}\left[w_{s,a} H_\mu(F)\vert_{(\xi, a)}\right] \right)^{-1}\left(\epsilon\cdot\mathbb{E}_{a^-\sim\pi^-(\cdot\vert s)}\left[ w_{s,a^-} \sup\vert \nabla_\mu F\vert \right]\right),
\end{equation*}
where $\sup|\nabla F|$ represents the supremum of the gradient magnitude of $F(\cdot)$ over the support range of $(\mu^+(s), a^-)$, $H(F)$ denotes the Hessian matrix of $F(\cdot)$, and $\xi=\vartheta\mu^+(s)+(1-\vartheta)\mu^*(s)$ with $\vartheta\in(0,1)$.
\end{theorem}
It can be seen that the upper bound of the deviation is jointly influenced by both the proportion $\epsilon$ of poor explorations in the offline data and the gradient upper bound $\sup|\nabla_\mu F|$. 
To better explain this bound, we may substitute the Gaussian distribution expression into Equation (\ref{eq7}), since in practice, the policy distribution is commonly assumed to follow $\pi(\cdot\vert s)\sim \mathcal{N}(\mu(s), \sigma^2(s)I)$, where $I$ denotes the identity matrix. For analytical convenience, we fix $\sigma(s)=\sigma$ as a constant, then we have:
\begin{lemma}\label{lemma6}
Assume the policy distribution follows $\pi(\cdot|s)\sim \mathcal{N}(\mu(s), \sigma^2I)$. Then, the AWRs' policy optimization problem is equivalent to:
\begin{equation}\label{eq9}
\min_\mu \mathbb{E}_{s,a\sim D}\left[w_{s,a}\frac{\left \|a-\mu(s)\right \|^2_2}{2\sigma^2}\right],
\end{equation}
where $w_{s,a}:=\exp{\left[\frac{1}{\lambda} A^{\pi_\beta}(s,a)\right]}$.
\end{lemma}
Obviously, $F(\mu(s),a)=\frac{\left \|a-\mu(s)\right \|^2_2}{2\sigma^2}$, thus we can obtain the following corollary.

% this is a weighted least squares problem, where the weight $w_{s,a}$ is proportional to the advantage function $A^{\pi_\beta}(s,a)$, representing the relative importance of each sample. During optimization, the algorithm preferentially fits state-action pairs with higher advantage values, thereby focusing on good explorations.

\begin{corollary}[An upper bound of the approximation bias introduced by poor explorations in AWRs based on normal distribution]\label{cor1}
Assume the policy distribution follows $\pi(\cdot|s)\sim \mathcal{N}(\mu(s), \sigma^2I)$. Then, for state $s$, the absolute difference between $\mu^*$ and $\mu^+$ satisfies the following inequality:
\begin{equation}
\vert \mu^*(s)-\mu^+(s)\vert \le \frac{\epsilon\cdot\mathbb{E}_{a^-\sim\pi^-(\cdot\vert s)}\left[ w_{s,a^-}\vert a^- - \mu^+(s)\vert \right]}{\mathbb{E}_{a\sim D}[w_{s,a}]}.
\end{equation}
\end{corollary}
This theorem provides an upper bound on the bias introduced by poor explorations when the policy distribution is a normal distribution. The bound is determined by: (1) the proportion $\epsilon$ of poor explorations in the dataset, and (2) the deviation $|a^- - \mu^+(s)|$ between poor and good explorations. When the offline data contains few poor explorations ($\epsilon \ll 1$) or when they are behaviorally similar to good ones ($|a^- - \mu^+(s)| \to 0$), the bias remains limited. Conversely, $L_2$-norm-based AWRs become affected by poor explorations, only obtain limited policy improvement.

\begin{figure}[H]%
 \centering
 \includegraphics[width=0.5\textwidth]{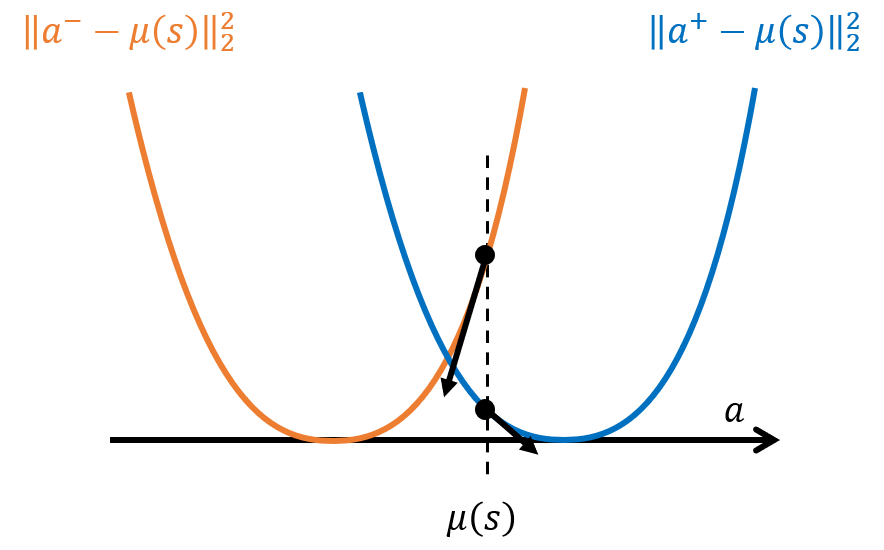}
 \caption{An overview of the gradient of the loss function based on the $L_2$ norm.}\label{fig1}
 \end{figure}

Indeed, the term $|a^- - \mu^+(s)|$ emerges from the derivative of the $L_2$ norm. From a gradient descent perspective, as $\mu(s)$ approaches $a^+$, the policy parameters become predominantly updated toward poor explorations as Fig. \ref{fig1} shows, driving convergence to suboptimal solutions. Therefore, the $L_2$-norm-based loss function may be particularly sensitive to poor explorations, thereby causing the over-conservatism problem.

\section{Corruption-Averse Advantage-Weighted Regression}

Building upon the theoretical analysis in the previous section, we identify two critical factors contributing to the over-conservatism problem in AWRs: (1) the sensitivity of the approximation loss function to poor explorations, and (2) the proportion of poor explorations in the offline data. To address this issue, we propose our improvements focusing on these two aspects: (1) using a more robust loss function that is less sensitive to poor explorations, and (2) designing a resampling method to reduce poor explorations in the training data, which we name as Advantage-based prioritized experience replay.

\subsection{Robust loss functions for approximation}
 
To design a more robust loss function, we introduce a $f(\cdot)$ to replace the $F(\cdot)$ in Equation (\ref{eq7}), i.e.
\begin{equation}\label{eq10}
\min_\mu \mathbb{E}_{s,a\sim D}\left[w_{s,a}\sum_{i=1}^df(a_i-\mu_i(s))\right].
\end{equation}
Since $\sup\vert \nabla_{\mu_j} \sum_{i=1}^df(a_i-\mu_i(s))\vert = \sup\vert \nabla_{\mu_j} f(a_j-\mu_j(s))\vert$, we only have to consider functions that have smaller gradient magnitude over the support range of $(a^- - \mu^+(s))$. Consequently, we define a family of loss functions $f(\cdot)$ whose gradient upper bounds are strictly smaller than that of the $L_2$ norm when $\vert a-\mu(s) \vert$ is large:
\begin{align*}
\text{L1}:\quad f(a-\mu(s)) &= \vert a-\mu(s) \vert,\\
\text{Huber \cite{Huber}}:\quad f(a-\mu(s)) &= \begin{cases}
 (a-\mu(s))^2 & \text{ if } \vert a-\mu(s) \vert\le \kappa \\
 2\kappa\vert a-\mu(s) \vert-\kappa^2 & \text{ if } \vert a-\mu(s) \vert > \kappa
\end{cases},\\
\text{Flat}:\quad f(a-\mu(s)) &= -\log\left[c_2\exp\left[-c_1(a-\mu(s))^2\right] + c_3\right]+c_4, \\
\text{Skew}:\quad f(a-\mu(s)) &= -\log\left[c_2\left(\exp\left[-c_1(a-\mu(s))^2\right] + \frac{1}{c_3\vert a-\mu(s) \vert+1}\right)\right]+c_4. 
\end{align*}

 \begin{figure}[H]%
 \centering
 \includegraphics[width=0.7\textwidth]{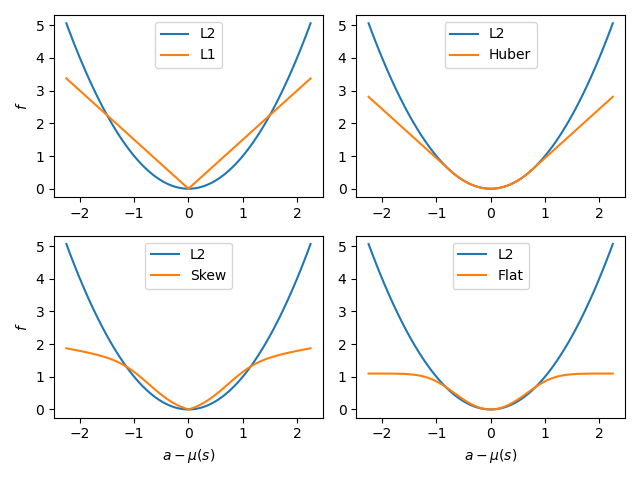}
 \caption{An overview of robust loss functions compared with $L_2$ norm.}\label{fig3}
 \end{figure}

Here $c_1, c_2, c_3\in \mathbb{R}^+$ are positive constants, with the Flat and Skew representing our newly proposed functions. Specifically, we construct the Flat function by introducing a constant term $c$ into the $L_2$ norm, yielding $-\log\left[\exp\left[-(a-\mu(s))^2\right]+c\right]$. Its derivative takes the form $\frac{-{\left[\exp\left[-(a-\mu(s))^2\right]\right]}' }{\exp\left[-(a-\mu(s))^2\right]+c}$, where when the policy function $\mu(s)$ deviates significantly from $a$, the exponential term $\exp\left[-(a-\mu(s))^2\right]$ approaches 0, causing the denominator to converge to the constant $c$. Consequently, the derivative asymptotically behaves as $-\frac{1}{c}{\left[\exp\left[-(a-\mu(s))^2\right]\right]}' $ and ultimately tends toward 0. To address potential oversuppression of gradients due to rapid convergence induced by the constant term, we further replace $c$ with a slowly decaying bounded function $\frac{1}{c_3\vert a-\mu(s) \vert+1}$, thereby creating the Skew function that is between the $L_1$ norm and the Flat function in terms of gradient behavior.

As visually compared to the $L_2$ norm in Fig. \ref{fig3}, these functions share the critical property of exhibiting smoother gradient profiles than the $L_2$ norm when $\mu(s)$ deviates significantly from $a$, thereby reducing algorithmic sensitivity to poor explorations during approximation. Specifically: (1) The $L_1$ norm assigns equal-magnitude but directionally distinct gradients for all $a\ne \mu(s)$; (2) The Huber function (Huber Loss \cite{Huber}), originally proposed by Huber in 1964, behaves as the $L_2$ norm when $\vert a-\mu(s) \vert\le \kappa$ and transitions to the $L_1$ norm for $\vert a-\mu(s) \vert> \kappa$; (3) The Skew function is between the $L_1$ norm and the Flat function, with progressively diminishing gradients as $\mu(s)$ moves away from $a$; (4) The Flat function asymptotically approaches zero gradient as $\vert a-\mu(s) \vert\to \infty$. 

Notably, while the Skew and Flat functions maintain convexity only near the origin (thus technically violating the prerequisites of Theorem \ref{thm2}), we strategically address this by initial hyperparameter tuning to confine all action samples within the convex region during early training phases, then gradually tightening this convex region as policy optimization progresses, marginalizing distant poor explorations beyond consideration.

\subsection{Advantage-based prioritized experience replay}

We build our resampling method based on Prioritized Experience Replay (PER) \cite{PER}, a reinforcement learning sampling method proposed by Schaul et al., which accelerates and improves the convergence of policy evaluation function estimation by prioritizing samples with higher Temporal-Difference (TD) Error during each iteration. Similarly, we adopt this framework to reduce the ratio of poor explorations in training batch by using the advantage function value $A^{\pi_\beta}(s,a)$ to design priority that characterize whether a sample represents good exploration, along with a hybrid sampling scheme that employs different sampling distributions for policy evaluation and policy optimization processes.

\begin{figure}[H]%
 \centering
 \includegraphics[width=0.5\textwidth]{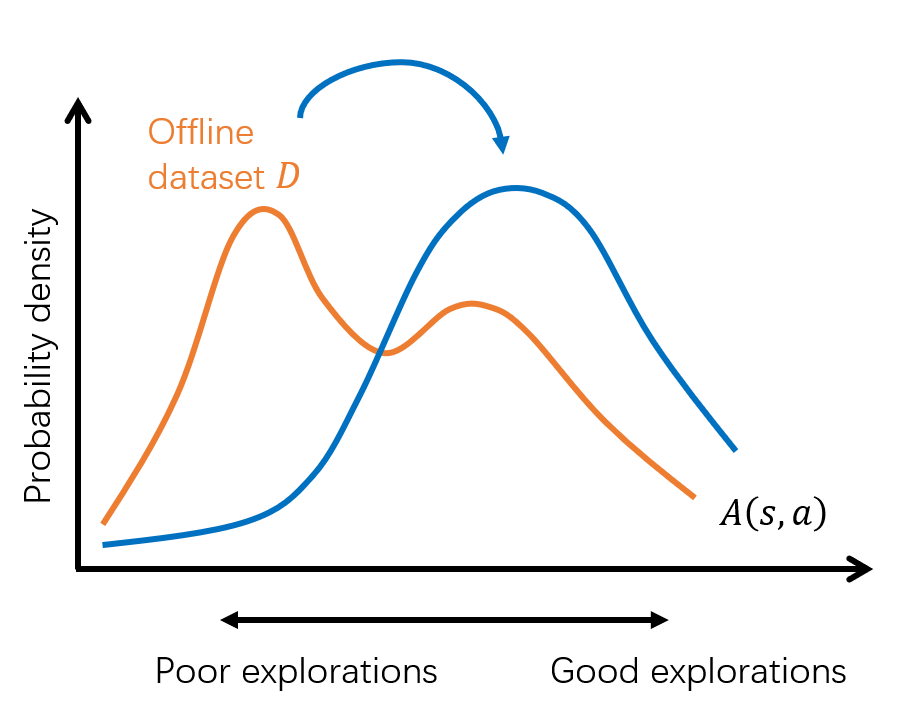}
 \caption{An overview of advantage-based prioritized experience replay.}\label{fig2}
 \end{figure}

For priority design, we set it as follows:
\begin{equation*}
p_i = h(A^{\pi_\beta}(s,a)), \forall s, a\in D,
\end{equation*}
where $h(\cdot)$ is a non-negative monotonically increasing function that we design in three variants: exponential, linear, and softmax types. Obviously,  samples with larger advantage function values $A^{\pi_\beta}(s,a)$ receive correspondingly higher sampling priorities, letting the proportion of good explorations become significantly enhanced in the training batch, see Fig. \ref{fig2}.

For exponential type, we have the following function form:
\begin{equation}\label{eq8}
p_i = \exp{\left[c_2 \left[A^{\pi_\beta}(s_i,a_i)-c_1\right]\right]},
\end{equation}
where $c_1$ and $c_2$ are two self-designed parameters that can be set as follows:
\begin{align*}
\text{Normal}:\quad c_1&=\hat{\mu}_A=\frac{1}{n}\sum_i A^{\pi_\beta}(s_i,a_i),\ c_2=\frac{1}{\lambda\cdot \hat{\sigma}_A},\\
\text{Quantile}:\quad c_1&=\hat{\tau}_A=\text{Quantile}(A^{\pi_\beta}(s_i,a_i)),\ c_2=\frac{1}{\lambda\cdot \hat{\sigma}_A},\\
\text{Standard}:\quad c_1&=0,\ c_2=\frac{1}{\lambda}.
\end{align*}
Indeed, since Lemma \ref{lemma1} establishes that the optimal policy based on KL divergence constraints takes the form $\pi^*_\beta(a|s) \propto \pi_\beta(a|s)\exp\left[\frac{1}{\lambda}A^{\pi_\beta}(s,a)\right]$, the distribution of training data will more closely approach this theoretically optimal policy distribution after employing exponential-form prioritized resampling.

For the other two priority types, linear and softmax, we set them as the priority defined in ODPR \cite{OPER}, along with the priority similar to the Advantage-Weighted (AW) method \cite{AW} (here extended from trajectory-level advantages to sample-level advantages):
\begin{align*}
\text{ODPR}:\quad p_i &= c\cdot [A^{\pi_\beta}(s_i,a_i)-\min_{s,a\in D}A^{\pi_\beta}(s,a)],\\
\text{AW}:\quad p_i &= \frac{\exp{\left[\frac{1}{\lambda} A^{\pi_\beta}(s_i,a_i)\right]}}{\sum_j\exp{\left[\frac{1}{\lambda} A^{\pi_\beta}(s_j,a_j)\right]}}.
\end{align*}

Furthermore, we aim to keep the algorithm simple without pretraining the $A^{\pi_\beta}(s,a)$ and updating the priorities during the iterative process. This approach raises two considerations: (1) the estimates of the advantage function $A^{\pi_\beta}(s,a)$ may contain substantial bias during early training stages, potentially affecting resampling accuracy and subsequent learning; (2) the estimation of policy evaluation functions $Q^{\pi_\beta}(s,a)$ and $V^{\pi_\beta}(s)$ requires samples that adequately cover the entire dataset $D$ to avoid introducing unnecessary bias. To address these concerns, we employ the sampling framework similar to Offline Decoupled Prioritized Resampling (ODPR) \cite{OPER} that utilizes both uniform sampling and advantage-based prioritized experience replay to generate two distinct equivalent sample batches, where the uniformly sampled batch is used for policy evaluation function estimation, and the prioritized batch focuses on policy optimization. The difference is that the priorities of both batches would be computed and updated later together, without using the new priority to multiply the old one iteratively.

\section{Theoretical analysis of CAWR}

By integrating the two aforementioned improvements - optimal policy approximation using robust loss functions and advantage-based prioritized experience replay, we arrive at our proposed CAWR, which can be formally expressed through the following optimization objective:
\begin{equation}\label{eq11}
\min_\pi\ \mathbb{E}_{s,a\sim \tilde{D}}\left[w_{s,a}\sum_{i=1}^df(a_i, \mu_i(s))\right],
\end{equation}
where $\tilde{D}$ denotes the sampled dataset after the resampling process. Although Theorem \ref{thm2} has provided evidence for the robustness of our proposed loss functions, we still need to know whether the new theoretical optimal policy after using advantage-based prioritized experience replay is better than the original. 

To formalize this, we define $\hat{M}=(\hat{\mathcal{S}}, \hat{\mathcal{A}}, \hat{p}, \hat{r}, \gamma)$ as the empirical Markov Decision Process (MDP) derived from dataset $D$, and let $\pi_{re}$ denote the behavior policy corresponding to the prioritized experience replay dataset $\tilde{D}$ in $\hat{M}$. Consequently, after applying prioritized experience replay, our algorithm is equivalent to solving the following optimization problem:
\begin{align*}
    \max_\pi\ J(\pi,\hat{M})-\lambda \mathbb{E}_{s\sim \tilde{D}} \left[D_{KL}(\pi\vert \vert \pi_{re})(s)\right].
\end{align*}
where $J(\pi,\hat{M})$ represents the return of policy $\pi$ in the empirical MDP $\hat{M}$. This leads us to the following lemma:
\begin{lemma}\label{lemma5}
Equation (\ref{eq11}) is equivalent to solving the following optimization problem:
\begin{align}\label{eq12}
    \max_\pi J(\pi,\hat{M})-\lambda \mathbb{E}_{s\sim \tilde{D}} \left[D_{KL}(\pi\vert \vert \pi_\beta)(s)+\mathbb{E}_{a\sim \pi,a\in D}\log \left(\frac{1}{h(A^{\pi_\beta}(s,a))}\right)\right].
\end{align}
\end{lemma}
Obviously, the policy constraint can be decomposed into two components: the first term $D_{KL}(\pi|\pi_\beta)(s)$ represents the standard policy constraint in AWRs, while the second term $\mathbb{E}_{a\sim \pi,a\in \mathcal{D}}\log \left(\frac{1}{h(A^{\pi_\beta}(s,a))}\right)$ is introduced by advantage-based prioritized experience replay. During optimization, for poor explorations where $A^{\pi_\beta}(s,a)$ is small, $\log \left(\frac{1}{h(A^{\pi_\beta}(s,a))}\right)$ becomes large, thereby counteracting part of the constraint from the first term; conversely, the constraint is strengthened for good explorations. Notably, when $h(\cdot)\equiv 1$, this formulation reduces to standard AWRs, demonstrating that AWRs constitute a special case of our algorithm. This leads to Theorem \ref{thm2}, which establishes that our theoretical optimal policy $\pi_{re}^*$ outperforms $\pi_\beta^*$.
\begin{theorem}\label{thm3}
Denote $\pi_{re}^*$ as the solution of Equation (\ref{eq11}), then
\begin{equation*}
\max_{\lambda, h}J(\pi_{re}^*;\lambda, h)\ge \max_{\lambda}J(\pi_\beta^*;\lambda).
\end{equation*}
\end{theorem}

\subsection{Practical implementation and algorithm}

The overall framework of our algorithm is illustrated in Algorithm \ref{algo1}. As shown, our implementation follows the IQL configuration \cite{IQL}, with bold blue text demonstrating our proposed additions to IQL for clear identification. For the policy distribution assumption, we similarly employ a fixed-variance normal distribution $\pi(\cdot|s) \sim \mathcal{N}(\mu(s), \sigma^2I)$, with the original $L_2$ norm replaced with our proposed loss functions when optimizing the policy. Furthermore, we omit the importance sampling (IS) weight correction commonly used in PER \cite{PER} to adjust for sample distribution shift.
% Furthermore, the overall framework of our algorithm is illustrated in Fig. \ref{fig4}. 
% During each parameter iteration, the algorithm samples two distinct training batches from the offline dataset $D$: batch $D_1$ obtained through uniform sampling and batch $D_2$ acquired via advantage-based prioritized experience replay, maintaining a 1:1 ratio between them. The algorithm then utilizes $D_1$ to estimate policy evaluation functions, while $D_2$ serves to solve the optimal policy through the optimization problem specified in Equation (\ref{eq10}) (employing any of the loss functions introduced previously: $L_p$ norms, Huber loss, Skew function, or Flat function). Notably, when using the $L_2$ norm, this formulation reduces to the policy iteration term in IQL \cite{IQL}. Finally, the algorithm merges $D_1$ and $D_2$ to recompute priorities (using any predefined priority schemes: Normal, Quantile, Standard, OPER, or AW priorities) and update these priorities in the sample pool.
 
%  \begin{figure}[H]%
%  \centering
%  \includegraphics[width=0.85\textwidth]{image/8.png}
%  \caption{An overview of Corruption-Averse Advantage-Weighted Regression (CAWR).}\label{fig4}
%  \end{figure}

 \begin{algorithm}[H]
\caption{Corruption-Averse Advantage-Weighted Regression (CAWR)}\label{algo1}
\begin{algorithmic}[1]
\Require Initialize $Q_{\theta_Q}(s,a)$, $V_{\theta_V}(s)$, $\mu_\phi(s)$. Input offline dataset $D$. Set Hyperparameters $\sigma,\ \lambda,\ \tau,\ w_{\max}, N, n$. 
\State $p_i\leftarrow 1, i=1,...,\vert D\vert$
\For{k$=0$ to $N$}
    \State $Q_k(s,a) \leftarrow Q_{\theta_Q}(s,a)$
    \State $V_k(s) \leftarrow V_{\theta_V}(s)$
    \State \textcolor{blue}{\textbf{Draw $n$ samples uniformly: $D_1=\{(s_i, a_i, r_i, s'_i)\}_{i=1}^n$.}}
	\State \textcolor{blue}{\textbf{Draw $n$ samples base on priorities: $D_2=\{(s_i, a_i, r_i, s'_i)\}_{i=1}^n$.}}
    \State \textcolor{blue}{\textbf{Use $D_1$ to compute}} the loss function of $V_{\theta_V}(s)$: $L(\theta_V)=\frac{1}{n} \sum _{i} L^\tau_2(Q_k(s_i,a_i) - V_{\theta_V}(s_i))$.
    \State Minimize the loss function $L(\theta_V)$, update $\theta_V$.
    \State \textcolor{blue}{\textbf{Use $D_1$ to compute}} the loss function of $Q_{\theta_Q}(s,a)$: $L(\theta_Q)=\frac{1}{n} \sum _{i} (r_i+\gamma V_k(s'_i) - Q_{\theta_Q}(s_i,a_i))^2$.
    \State Minimize the loss function $L(\theta_Q)$, update $\theta_Q$.
    \State $A_k(s,a) \leftarrow Q_{\theta_Q}(s,a) - V_{\theta_V}(s)$
    \State For all samples in $D_1\cup D_2$, compute $w_i \leftarrow \min \left(\exp{\left[c_2 (A_k(s_i,a_i)-c_1)\right]}, w_{\max}\right)$.
    \State \textcolor{blue}{\textbf{Use $D_2$ to compute the loss function of $\mu_\phi(s)$ (see (\ref{eq10})): $J(\phi) = \frac{1}{n}\sum_i \left[\frac{w_i}{2\sigma^2}\sum_{j=1}^df\left(a_{i,j}-\mu_{\phi,j}(s_i)\right)\right]$.}}
    \State Minimize the loss function $J(\phi)$, update $\phi$.
	\State \textcolor{blue}{\textbf{For all samples in $D_1\cup D_2$, update priorities $p_i \leftarrow h(A_k(s_i,a_i))$.}}
\EndFor
\end{algorithmic}
\end{algorithm}

\section{Experiments}
% \subsection{Experiment settings}

We evaluate our algorithm on the D4RL benchmark and compare it with IQL, one of the state-of-the-art offline reinforcement learning algorithms. The algorithm's performance is evaluated using the normalized episode return of the policy, where a score approaching 100 indicates near-expert capability, while a score near 0 represents near-random performance. To evaluate the performance of the baseline algorithm, we utilize results directly from the original papers that proposed IQL \cite{IQL}.  For the $k$-th parameter iteration, we define $\text{score}_k$ as the highest score achieved up to that point: $\text{score}_k = \max_{i=0,...,k} \text{score}(\pi_i)$.

The D4RL benchmark includes diverse RL task datasets, and our experiments mainly focus on the Mujoco locomotion tasks: Hopper, Walker2D, and HalfCheetah, which originate from simulation environments in the Gymnasium library \cite{Gymnasium}, rendered by the MuJoCo physics engine \cite{mujoco}, and the objective is to control the robot's torso to go quickly and stably. For each task, the dataset contains five data types: random, medium, medium-replay, medium-expert, and expert. Our study mainly utilizes the medium, medium-expert, and expert datasets: (1) The medium dataset consists of trajectories collected by a medium-level policy trained via SAC \cite{SAC1, SAC2}; (2) The medium-expert dataset contains an equal mixture of data from both medium-level and expert-level policies, both trained with SAC; (3) The expert dataset comprises trajectories generated by an expert-level SAC-trained policy.

Meanwhile, $L_2$-norm-based CAWR is denoted as L2, which essentially replicates the standard IQL algorithm. L1, Huber, Skew, and Flat represent variants employing the $L_1$ norm, Huber loss, Skew function, and Flat function, respectively. We directly use the Laplace distribution as the policy distribution since its corresponding $F(\cdot)$ is the $L_1$ norm. All resampling methods are collectively referred to as PER, with the specific priority distinguished through Normal, Standard, Quantile, AW, and ODPR. More details are provided in Appendix \ref{app0}.

\subsection{Ablation analysis}

Here, we first pre-train an advantage function $A_{\tau}(s,a)$ and then perform policy optimization exclusively during testing. This aims to eliminate interference caused by the dynamic updates of the advantage function during training, thereby precisely evaluating the impact of different loss functions (i.e., L2, L1, Huber, Skew, Flat) and priorities (i.e., Normal, Quantile, AW, ODPR) on policy optimization.
The results of the ablation study are presented in Table \ref{tab1}, the figures that visualize the score's progression across iterations can be found in Appendix \ref{app1}. 

\textbf{Loss function evaluation:} As the loss function transitions from L2 to the Flat function, the gradient magnitude at points distant from zero decays, thereby reducing sensitivity to poor explorations and ultimately yielding superior learned policy. Among the four proposed loss functions, L1 and Flat perform best, while Huber loss bears the closest resemblance to L2 — it behaves quadratically near zero (like L2) while adopting linear characteristics (like L1) farther from the origin. This hybrid nature explains why its score progression most closely mirrors that of L2, indicating that its sensitivity to poor explorations ranks second only to L2 itself.

\textbf{Priority evaluation:} Regardless of the choice of loss function or priority definition, incorporating Advantage-based PER yields significant improvements in scores compared to 'None' implementations. The improvement of different priorities varies across loss functions: Normal and Median (Quantile $\tau$=0.5) demonstrate the most substantial enhancements, achieving scores above 80 in three out of the four loss functions. Standard follows closely, attaining scores exceeding 80 in two loss function configurations, with the Flat function variant reaching 96 points, nearing expert-level performance. Other priority definitions show comparatively modest improvements. These results indicate that priorities differentially affect the sampling efficiency of good explorations, suggesting that proper priority selection can dramatically boost the score for a specific loss function.

\begin{table}[H]
\begin{center}
\begin{minipage}{\textwidth}
\caption{The results of the ablation study, where 'None' indicates the absence of PER and bold values highlight the top-2 scores within each experimental group.}\label{tab1}%
\resizebox{1.1\linewidth}{!}{
\begin{tabular}{lrr|lrrrrr}
\toprule
Dataset & BC (repro.) & \makecell[r]{IQL\\(paper)} & PER Type  & L2 & L1 & Huber & Skew & Flat \\
\midrule
\multirow{7}*{hopper-m-v2} & $61.80\pm11.91$ & 66.3 & None & $59.05\pm11.47$ & $61.26\pm12.04$ & $59.09\pm7.79$ & $\textbf{70.84}\pm6.50$ & $\textbf{76.23}\pm8.82$  \\
~ & $61.80\pm11.91$ & 66.3 & Normal & $\textbf{84.42}\pm18.32$ & $\textbf{87.10}\pm17.54$ & - & $82.92\pm20.58$ & $47.26\pm1.70$  \\
~ & $61.80\pm11.91$ & 66.3 & Standard & $77.49\pm17.72$ & $\textbf{80.75}\pm19.89$ & - & $68.36\pm19.26$ & $\textbf{96.00}\pm7.24$  \\
~ & $61.80\pm11.91$ & 66.3 & AW & $67.42\pm15.00$ & $71.82\pm17.52$ & - & $\textbf{74.83}\pm12.26$ & $\textbf{79.81}\pm10.00$  \\
~ & $61.80\pm11.91$ & \textbf{66.3} & ODPR & $63.91\pm7.78$ & $63.91\pm9.82$ & - & $66.22\pm13.92$ & $\textbf{83.63}\pm8.56$  \\
~ & $61.80\pm11.91$ & 66.3 & Quantile $\tau=0.5$ & $\textbf{89.00}\pm15.28$ & $\textbf{83.07}\pm20.72$ & - & $82.52\pm19.06$ & $51.70\pm18.23$  \\
~ & $61.80\pm11.91$ & 66.3 & Quantile $\tau=0.3$ & $78.42\pm20.26$ & $\textbf{82.22}\pm20.53$ & - & $\textbf{78.95}\pm21.89$ & $30.10\pm2.14$  \\
\bottomrule
\end{tabular}
}
\end{minipage}
\end{center}
\end{table}

\subsection{Comparison on the D4RL benchmarks}

We evaluate the complete Algorithm \ref{algo1} (L1 loss function with Normal PER) without pre-training the advantage function. The results are presented in Table \ref{tab2}, while the progression of scores across iterations for different robotic tasks is detailed in Appendix \ref{app1}.

The experimental results show that our algorithm (L1+PER(Normal)) achieves the best overall performance, ranking top-2 in 7 out of 9 datasets, followed by the solely-Normal-PER variant (L2+PER(Normal)). While L1 generally outperforms L2 regardless of PER usage across most scenarios, this result reverses specifically for Walker2D's medium-expert and expert-level datasets, where L2 shows better scores. The Normal PER enhancement proves significantly beneficial in 7 of 9 datasets for both L1 and L2 configurations, with the exceptions being Hopper-expert and HalfCheetah-expert datasets. Notably, BC achieves second-best performance on both expert-level datasets, suggesting that direct policy imitation proves more effective than PER-based optimization, likely because the sampled trajectories from these expert datasets may not reliably estimate the true optimal policy.

\begin{table}[H]
\begin{center}
\begin{minipage}{\textwidth}
\caption{The results of performance comparison, where 'm' and 'e' are the abbreviations of 'medium' and 'expert', and bold values highlight the top-2 scores within each experimental group.}\label{tab2}%
\resizebox{\linewidth}{!}{
\begin{tabular}{lrr|rr|rr}
\toprule
Dataset & BC (repro.) & \makecell[r]{IQL (paper)} & \makecell[r]{IQL (repro.)\\L2} & \makecell[r]{CAWR\\L1} & \makecell[r]{CAWR\\L2+PER(Normal)} & \makecell[r]{CAWR\\L1+PER(Normal)} \\
\midrule
hopper-m-v2 & $61.80\pm11.91$ & 66.3 & $60.40\pm7.68$ & $61.09\pm13.43$ & $\textbf{70.68}\pm19.44$ & $\textbf{85.52}\pm17.48$  \\
hopper-m-e-v2 & $76.46\pm20.27$ & 91.5 & $78.88\pm21.10$ & $81.17\pm20.69$ & $\textbf{109.92}\pm11.75$ & $\textbf{111.05}\pm4.73$  \\
hopper-e-v2 & $\textbf{111.47}\pm0.57$ & - & $111.32\pm1.17$ & $\textbf{112.21}\pm1.17$ & $110.47\pm5.23$ & $111.34\pm5.19$  \\
\midrule
walker2d-m-v2 & $81.01\pm1.92$ & 78.3 & $80.23\pm5.74$ & $81.02\pm2.83$ & $\textbf{83.35}\pm2.23$ & $\textbf{84.75}\pm3.00$  \\
walker2d-m-e-v2 & $109.75\pm0.34$ & 109.6 & $109.48\pm0.24$ & $109.25\pm0.21$ & $\textbf{111.21}\pm0.51$ & $\textbf{110.90}\pm0.43$  \\
walker2d-e-v2 & $109.62\pm0.18$ & - & $109.74\pm0.29$ & $109.03\pm0.57$ & $\textbf{111.86}\pm0.47$ & $\textbf{111.10}\pm0.61$  \\
\midrule
halfcheetah-m-v2 & $43.07\pm0.84$ & \textbf{47.4} & $43.31\pm1.05$ & $43.51\pm1.23$ & $45.69\pm0.49$ & $\textbf{45.87}\pm0.83$  \\
halfcheetah-m-e-v2 & $61.97\pm21.34$ & 86.7 & $62.58\pm21.90$ & $59.12\pm17.58$ & $\textbf{92.56}\pm2.77$ & $\textbf{93.30}\pm2.79$  \\
halfcheetah-e-v2 & $\textbf{94.15}\pm1.08$ & - & $94.04\pm 0.68$ & $\textbf{95.26}\pm0.72$ & $93.37\pm1.68$ & $90.89\pm10.67$  \\
\bottomrule
\end{tabular}
}
\end{minipage}
\end{center}
\end{table}

\section{Related works}
\subsection{Constrain-based offline RL}
Constrained offline reinforcement learning algorithms primarily address the distribution shift problem by introducing constraints or penalty terms into the optimization process. For instance, Fujimoto et al. proposed a batch-constrained policy to reduce extrapolation errors introduced by distribution shift. Several works have incorporated explicit policy constraints: the TD3+BC algorithm \cite{TD3_BC} introduces a behavior cloning term \cite{BC} to constrain the policy; the Bootstrapping Error Accumulation Reduction (BEAR) \cite{BEAR} employs Maximum Mean Discrepancy (MMD) to penalize differences between policy distributions; AWRs utilize KL divergence as the policy constraint penalty; and the Behavior Proximal Policy Optimization (BPPO) \cite{BPPO} adopts the clipping in Proximal Policy Optimization (PPO) \cite{PPO} to enforce policy constraints. Additionally, value regularization methods implicitly constrain policies by adjusting Q-function values for in-distribution versus out-of-distribution data. For example, CQL \cite{CQL} modifies Q function to assign lower values to out-of-distribution actions, while Mildly Conservative Q-Learning (MCB) \cite{MCB} refines CQL's regularization approach to enhance generalization. Beyond these, An et al. \cite{EDAC} and Wu et al. \cite{UWAC} incorporate uncertainty-based penalties into the Q function, whereas Kidambi et al. \cite{MOReL} and Yu et al. \cite{Mopo} further fit an environment dynamic model to penalize out-of-distribution regions, thus ensuring safer policy optimization.

\subsection{Prioritized dataset}
Several studies have employed resampling techniques to enhance the proportion of good explorations in the dataset, thus making the agent focus more on good samples. For instance, Advantage-Weighting (AW)/Return-Weighting (RW) methods \cite{AW} and Percentage-Filtering (PF) approach \cite{PF} utilize trajectory returns or advantage values to filter high-quality trajectory data. The Offline Decoupled Prioritized Replay (ODPR) method \cite{OPER} adopts advantage-based/return-based iterative priority for decoupled prioritized resampling. The Density-ratio Weighting (DW) technique \cite{DW} implements sample weighting, where the computed weights achieve a balance between maximizing weighted total returns and minimizing distributional shift.

\section{Conclusion}
Our research mainly investigates the over-conservatism issue in AWRs. Through theoretical analysis, we identify two critical contributing factors: (1) the sensitivity of approximation loss functions to poor explorations, and (2) the proportion of such poor explorations in offline data. To address this, we propose our CAWR algorithm, which incorporates two key innovations: a group of more robust loss functions designed to reduce sensitivity to poor explorations, and an advantage-based prioritized experience replay for filtering poor explorations out. Numerical evaluations on the D4RL benchmark demonstrate that CAWR effectively mitigates over-conservatism in AWRs, enabling superior policy learning from suboptimal offline datasets.

\section*{Acknowledgments}
The compute cluster for the experiments was provided by the Institute of Science and Technology for Brain-Inspired Intelligence (ISTBI).

%Bibliography
\bibliographystyle{unsrt}  
\bibliography{references}  

\newpage
\begin{appendix}

\section{Proof of lemmas and theorems}
\subsection{Proof of Lemma \ref{lemma0}}
\begin{proof}
The proof follows a similar derivation process to Lemma \ref{lemma1}. 

First, Equation (\ref{eq6}) can be further expanded as:
\begin{align*}
    \mathcal{L}(\pi, \lambda) &= \mathbb{E}_{s\sim D, a\sim \pi(\cdot\vert s)}[A^{\pi_\beta}(s, a)] + \lambda\cdot \mathcal{H}(\pi)\\
    &= \mathbb{E}_{s\sim D, a\sim \pi(\cdot\vert s)}\left[A^{\pi_\beta}(s, a) - \lambda \cdot \log \pi(a\vert s)\right]\\
    &= \mathbb{E}_{s\sim D, a\in \mathcal{A}}\left[\pi(a\vert s)A^{\pi_\beta}(s, a) - \lambda \cdot \pi(a\vert s)\log \pi(a\vert s)\right],
\end{align*}
Next, we compute the partial derivative of this expression with respect to policy $\pi$:
\begin{align*}
    \nabla_\pi \mathcal{L}(\pi, \lambda) &= \mathbb{E}_{s\sim D, a\in \mathcal{A}}\left[A^{\pi_\beta}(s, a) - \lambda \cdot \left(\log \pi(a\vert s) + \pi(a\vert s)\frac{1}{\pi(a\vert s)}\right)\right]\\
    &=\mathbb{E}_{s\sim D, a\in \mathcal{A}}\left[A^{\pi_\beta}(s, a) - \lambda \cdot \left(\log \pi(a\vert s) + 1\right)\right],
\end{align*}
Under the constraint $\mathbb{E}{a\in \mathcal{A}}\pi^*(a|s)=1$, setting $\nabla\pi \mathcal{L}(\pi, \lambda)=0$, then the optimal policy $\forall s\sim \mathcal{D}$ is:
$$\pi^*(a\vert s) = \frac{1}{Z(s)}\exp{\left[\frac{1}{\lambda} A^{\pi_\beta}(s,a)\right]}.$$
\end{proof}

\subsection{Proof of Theorem \ref{thm1}}
\begin{proof}
Since
\begin{align}
    \mathcal{H}(\pi^*, \pi_\beta^*) &= \mathbb{E}_{a\sim \pi^*(\cdot\vert s)}[-\log \pi_\beta^*(a\vert s)]\nonumber\\
    &= \mathbb{E}_{a\sim \pi^*(\cdot\vert s)}\left[ -\log\pi_\beta(a\vert s)-\frac{1}{\lambda}A^{\pi_\beta}(s,a)+\log Z_\beta(s) \right]\nonumber\\
    &= \mathbb{E}_{a\sim \pi^*(\cdot\vert s)}[-\log\pi_\beta(a\vert s)]\nonumber\\
    &+\mathbb{E}_{a\sim \pi^*(\cdot\vert s)}\left[-\log \exp{\left[\frac{1}{\lambda}A^{\pi_\beta}(s,a)\right]}\right]\nonumber\\
    &+\log \sum_{a\in \mathcal{A}}\left[ \pi_\beta(a\vert s) \exp{\left(\frac{1}{\lambda} A^{\pi_\beta}(s,a)\right)}\right]\nonumber.
\end{align}
By Jensen's inequality for logarithmic functions, we have
\begin{align*}
    \log \sum_{a\in \mathcal{A}}\left[ \pi_\beta(a\vert s) \exp{\left[\frac{1}{\lambda} A^{\pi_\beta}(s,a)\right]}\right] 
    &\ge \mathbb{E}_{a\sim \pi_\beta(\cdot\vert s)}\left[\log \exp{\left[\frac{1}{\lambda} A^{\pi_\beta}(s,a)\right]}\right]\\
    &= \mathbb{E}_{a\sim \pi_\beta(\cdot\vert s)}\left[\log \pi^*(a\vert s)\right]+\log Z(s).
\end{align*}
Meanwhile,
\begin{align*}
    \mathbb{E}_{a\sim \pi^*(\cdot\vert s)}\left[-\log \exp{\left[\frac{1}{\lambda}A^{\pi_\beta}(s,a)\right]}\right] &= \mathbb{E}_{a\sim \pi^*(\cdot\vert s)}\left[-\log \pi^*(a\vert s)\right]-\log Z(s).
\end{align*}
Thus
\begin{align*}
    \mathcal{H}(\pi^*, \pi_\beta^*) 
    &\ge \mathbb{E}_{a\sim \pi^*(\cdot\vert s)}[-\log\pi_\beta(a\vert s)]\\
    &+\mathbb{E}_{a\sim \pi^*(\cdot\vert s)}\left[-\log \pi^*(a\vert s)\right]\\
    &-\mathbb{E}_{a\sim \pi_\beta(\cdot\vert s)}\left[-\log \pi^*(a\vert s)\right].
\end{align*}
By the relationship between cross-entropy and KL divergence, we have
\begin{equation*}
    D_{KL}(\pi^*\vert \vert\pi_\beta^*) \ge \mathcal{H}(\pi^*, \pi_\beta)-\mathcal{H}(\pi_\beta, \pi^*).
\end{equation*}
\end{proof}

\subsection{Proof of Theorem \ref{thm2}}
\begin{proof}
The proof follows a similar derivation process to paper \textit{Robust Estimation of a Location Parameter}\cite{Huber}. 

According to the optimality conditions for unconstrained convex optimization problems \cite{convex, convex2}, the global optimal solution $\mu^*$ of Equation (\ref{eq7}) satisfies
\begin{equation}\label{eq13}
(1-\epsilon)\cdot\mathbb{E}_{a^+\sim\pi^+(\cdot\vert s)}\left[w_{s,a^+}\nabla _\mu F\left (\mu^*(s), a^+\right )\right] + \epsilon\cdot\mathbb{E}_{a^-\sim\pi^-(\cdot\vert s)}\left[w_{s,a^-}\nabla _\mu F\left (\mu^*(s), a^-\right )\right]=0,
\end{equation}
for all $s\in \mathcal{S}$.

By performing a Taylor expansion of the left-hand side function in Equation (\ref{eq13}) around $\mu^+$ with $\mu^*$ as the variable, we obtain
\begin{equation*}
\epsilon\cdot\mathbb{E}_{a^-\sim\pi^-(\cdot\vert s)}\left[ w_{s,a^-}\nabla _\mu F\left (\mu^+(s), a^-\right )\right]+\mathbb{E}_{a\sim D}\left[w_{s,a} H_\mu(F)\vert_{(\xi, a)}\right]\left(\mu^*(s)-\mu^+(s)\right)=0.
\end{equation*}
By Jensen's inequality, we have
\begin{align*}
\vert \mu^*(s)-\mu^+(s)\vert &=\left\vert \left( \mathbb{E}_{a\sim D}\left[w_{s,a} H_\mu(F)\vert_{(\xi, a)}\right] \right)^{-1}\left(\epsilon\cdot\mathbb{E}_{a^-\sim\pi^-(\cdot\vert s)}\left[ w_{s,a^-} \nabla _\mu F\left (\mu^+(s), a^-\right ) \right]\right)\right\vert\\
&\le \left( \mathbb{E}_{a\sim D}\left[w_{s,a} H_\mu(F)\vert_{(\xi, a)}\right] \right)^{-1}\left(\epsilon\cdot\mathbb{E}_{a^-\sim\pi^-(\cdot\vert s)}\left[ w_{s,a^-} \sup\vert \nabla_\mu F\vert \right]\right).
\end{align*}
\end{proof}

\subsection{Proof of Lemma \ref{lemma6}}
\begin{proof}
Since
\begin{align*}
\pi(a \vert s) &= \frac{1}{(2\pi\sigma^2)^{d/2}}\exp\left[ -\frac{1}{2\sigma^2}\sum_{i=1}^d (a_i-\mu_i(s))^2\right].
\end{align*}
Thus
\begin{align*}
\log \pi(a\vert s)&\exp{\left[\frac{1}{\lambda} A^{\pi_\beta}(s,a)\right]} \\
&= \exp{\left[\frac{1}{\lambda} A^{\pi_\beta}(s,a)\right]}\left[ -\frac{1}{2\sigma^2}\sum_{i=1}^d (a_i-\mu_i(s))^2 - \frac{d}{2}\log (2\pi) - d\log(\sigma)\right]\\
&=  -\exp{\left[\frac{1}{\lambda} A^{\pi_\beta}(s,a)\right]}\frac{\sum_{i=1}^d (a_i-\mu_i(s))^2}{2\sigma^2} + c\\
&= -\frac{1}{2\sigma^2}\exp{\left[\frac{1}{\lambda} A^{\pi_\beta}(s,a)\right]}\left \|a-\mu(s)\right \|^2_2 + c
\end{align*}
\end{proof}

\subsection{Proof of Lemma \ref{lemma5}}
\begin{proof}
Since $\pi_{re}(a\vert s)= \frac{1}{Z(s)}\pi_\beta(a\vert s) h(A^{\pi_\beta}(s,a))$, we have
\begin{align*}
D_{KL}(\pi \vert\vert \pi_{\text{re}})(s) &= \mathbb{E}_{a\sim \pi, a\in D}\left[-\log\left( \frac{1}{Z(s)}\pi_\beta(a\vert s) h(A^{\pi_\beta}(s,a)) \right)\right]\\
&=\mathbb{E}_{a\sim \pi, a\in D}\left[-\log\pi_\beta(a\vert s)\right] + \mathbb{E}_{a\sim \pi, a\in D}\log\left( \frac{1}{h(A^{\pi_\beta}(s,a))} \right) + \mathbb{E}_{a\sim \pi, a\in D}[\log Z(s)].
\end{align*}
None that $\mathbb{E}_{a\sim \pi, a\in D}\left[-\log\pi_\beta(a\vert s)\right]=D_{KL}(\pi \vert\vert \pi_\beta)(s)$, and $\mathbb{E}_{a\sim \pi, a\in D}[\log Z(s)]$ has no relationship with $\pi$, we can drop it when solving the optimization problem.
\end{proof}

\newpage
\section{Experimental Details}\label{app0}

We utilize the DI-engine \cite{DI_engine} library as the foundational framework for our code implementation. The score of each policy is computed as follows:
\begin{equation*}
\text{score}(\pi):=\frac{J(\pi)-J(\pi_{r})}{J(\pi_{e})-J(\pi_{r})}\times 100,
\end{equation*}
where the values of $J(\pi_{r})$ (random policy) and $J(\pi_{e})$ (expert policy) for different tasks are provided in Table \ref{tab3}. 
\begin{table}[H]
\begin{center}
\begin{minipage}{0.35\textwidth}
\caption{The values of $J(\pi_{r})$ and $J(\pi_{e})$.}\label{tab3}%
\begin{tabular}{l r r}
\toprule
Task & $J(\pi_{r})$ & $J(\pi_{e})$ \\
\midrule
Hopper & -20.27 & 3234.3 \\
Walker2D & 1.63 & 4592.3 \\
HalfCheetah & -280.18 & 12135.0 \\
\bottomrule
\end{tabular}
\end{minipage}
\end{center}
\end{table}

For model training and evaluation, we conduct experiments with three random seeds, fixing one seed per training run with 400,000 iterations, and record the policy every 10,000 iterations. During evaluation, each trained policy interacts with the environment for 10 episodes to calculate the mean and standard deviation of the scores. These results are then aggregated across all random seeds at the same iteration point to compute the final mean $\text{score}_k$ and standard deviation. Since our IQL implementation is reproduced from the original paper, performance may differ slightly from the reported results. To ensure fair comparison, we use the original paper's scores as the baseline (indicated by a black dashed line in the results figures, see Appendix \ref{app1}).

The details of the hyperparameters are as follows (Table \ref{tab4} and Table \ref{tab5}).
\begin{table}[H]
\begin{center}
\begin{minipage}{0.4\textwidth}
\caption{Shared hyperparameter setup of CAWR (our algorithm) for diverse tasks.}\label{tab4}%
\begin{tabular}{l r r}
\toprule
Hyperparameters & CAWR \\
\midrule
Policy learning rate & 3e-4 \\
Value function learning rate & 3e-4 \\
Q function learning rate & 3e-4 \\
Soft target update coefficient & 0.005 \\
Batch size & 512 \\
Discount factor & 0.99 \\
initial log std of policy distribution & -2 \\
$\tau$ for IQL & 0.7 \\
\bottomrule
\end{tabular}
\end{minipage}
\end{center}
\end{table}

\begin{table}[H]
\begin{center}
\begin{minipage}{0.7\textwidth}
\caption{Hyperparameter setup of CAWR (our algorithm) for each task, where $\sigma$ is the std of the policy distribution.}\label{tab5}%
\begin{tabular}{l l r}
\toprule
Task & Hyperparameters & CAWR \\
\midrule
\multirow{5}*{hopper-m-v2} & $\lambda$ in Equation (\ref{eq5}) & $1/5$\\
~ & $w_{\max}$ for clipping weights and priorities & 10000\\
~ & $\kappa$ for Huber loss & 0.2 \\
~ & Flat $(c_1, c_2, c_3)$ & $(2/\sigma^2, 1/\sigma, 0.5)$\\
~ & Skew $(c_1, c_2, c_3)$ & $(1/\sigma^2, 1/\sigma, 1/\sigma)$\\
\multirow{2}*{hopper-m-e-v2} & $\lambda$ in Equation (\ref{eq5}) & $1/3$\\
~ & $w_{\max}$ for clipping weights and priorities & 100\\
\multirow{2}*{hopper-e-v2} & $\lambda$ in Equation (\ref{eq5}) & $1/3$\\
~ & $w_{\max}$ for clipping weights and priorities & 100\\
\midrule
\multirow{2}*{walker2d-m-v2} & $\lambda$ in Equation (\ref{eq5}) & $1/3$\\
~ & $w_{\max}$ for clipping weights and priorities & 10000\\
\multirow{2}*{walker2d-m-e-v2} & $\lambda$ in Equation (\ref{eq5}) & $1/3$\\
~ & $w_{\max}$ for clipping weights and priorities & 100\\
\multirow{2}*{walker2d-e-v2} & $\lambda$ in Equation (\ref{eq5}) & $1/3$\\
~ & $w_{\max}$ for clipping weights and priorities & 100\\
\midrule
\multirow{2}*{halfcheetah-m-v2} & $\lambda$ in Equation (\ref{eq5}) & $1/3$\\
~ & $w_{\max}$ for clipping weights and priorities & 10000\\
\multirow{2}*{halfcheetah-m-e-v2} & $\lambda$ in Equation (\ref{eq5}) & $1/3$\\
~ & $w_{\max}$ for clipping weights and priorities & 100\\
\multirow{2}*{halfcheetah-e-v2} & $\lambda$ in Equation (\ref{eq5}) & $1/3$\\
~ & $w_{\max}$ for clipping weights and priorities & 100\\
\bottomrule
\end{tabular}
\end{minipage}
\end{center}
\end{table}

\section{Supplementary results}\label{app1}

 \begin{figure}[H]%
 \centering
 \includegraphics[width=0.8\textwidth]{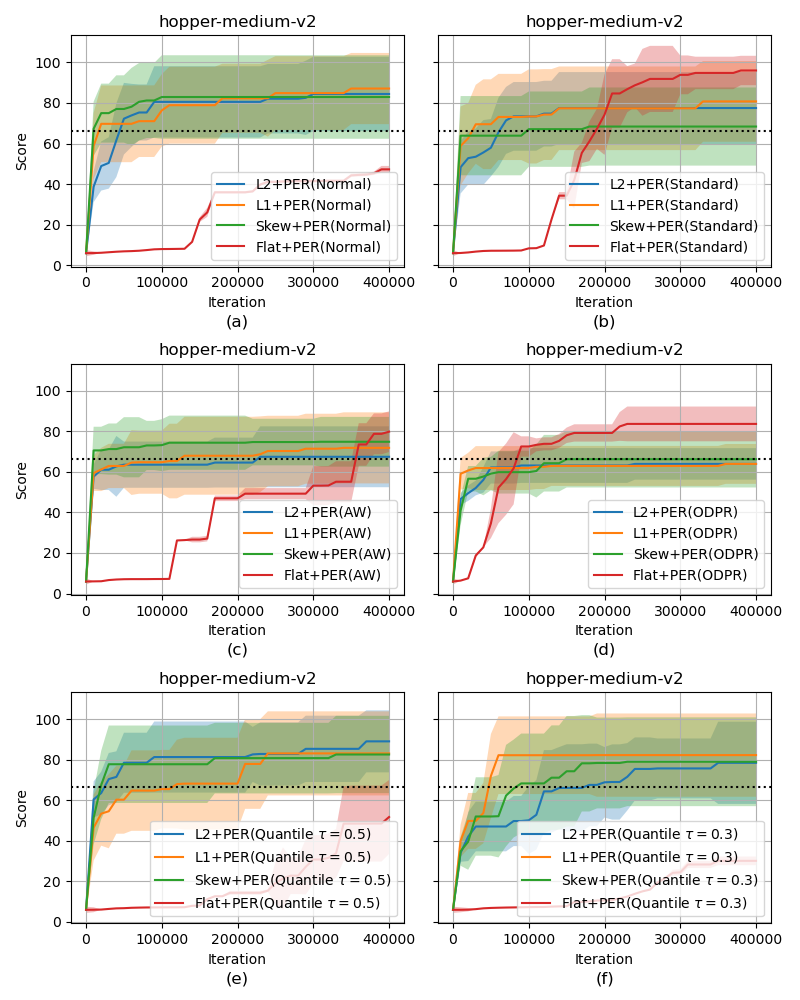}
 \caption{The scores across iterations for policy optimization using different priorities, where the shaded regions represent the score's mean $\pm$ one standard deviation, the black dotted line indicates the score of IQL (paper).}\label{fig5}
 \end{figure}
 
\begin{figure}[H]%
\centering
\includegraphics[width=0.8\textwidth]{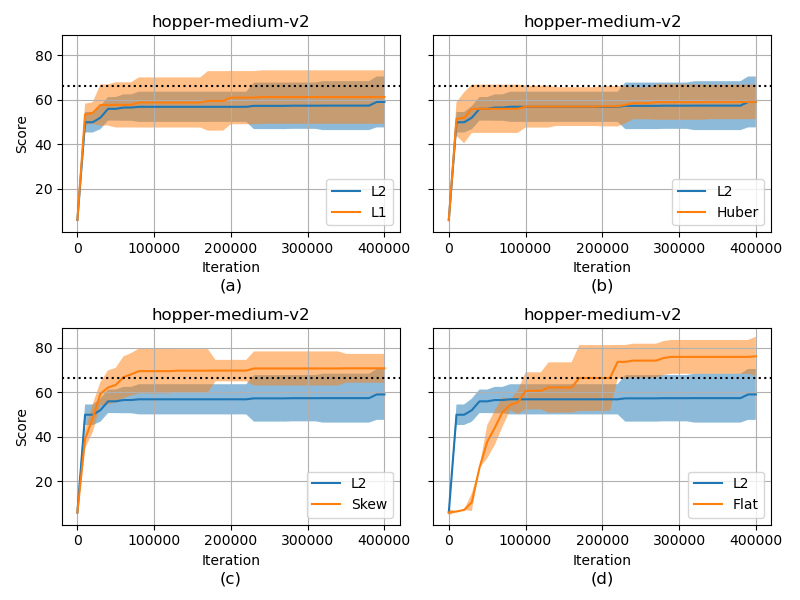}
\caption{The scores across iterations for policy optimization using different loss functions, where the shaded regions represent the score's mean $\pm$ one standard deviation, the black dotted line indicates the score of IQL (paper).}\label{fig4}
\end{figure}

 \begin{figure}[H]%
 \centering
 \includegraphics[width=0.8\textwidth]{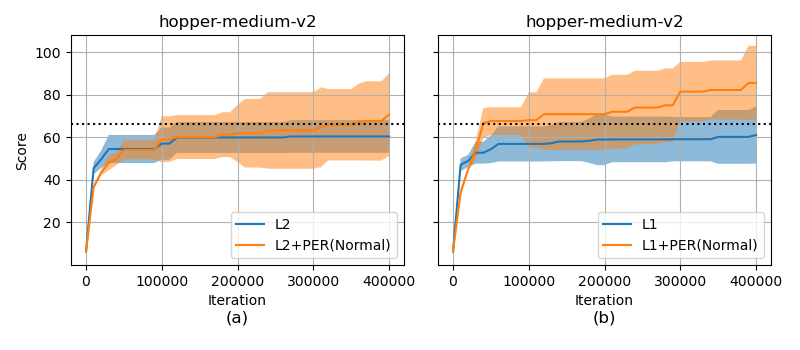}
 \caption{The scores across iterations for policy optimization tested on hopper-median dataset, where the shaded regions represent the score's mean $\pm$ one standard deviation, the black dotted line indicates the score of IQL (paper).}\label{fig6}
 \end{figure}
 \begin{figure}[H]%
 \centering
 \includegraphics[width=0.8\textwidth]{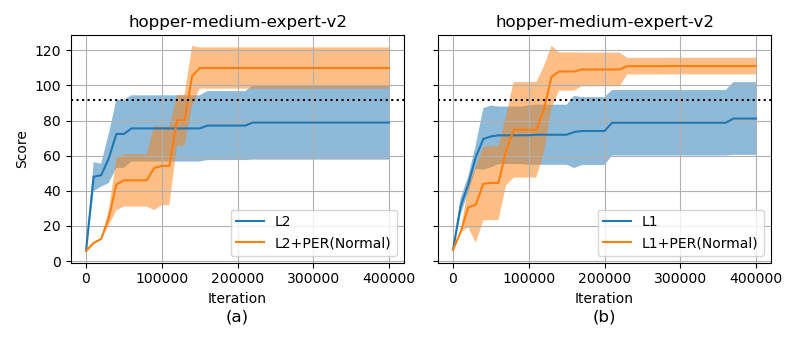}
 \caption{The scores across iterations for policy optimization tested on hopper-medium-expert dataset, where the shaded regions represent the score's mean $\pm$ one standard deviation, the black dotted line indicates the score of IQL (paper).}\label{fig7}
 \end{figure}
 \begin{figure}[H]%
 \centering
 \includegraphics[width=0.8\textwidth]{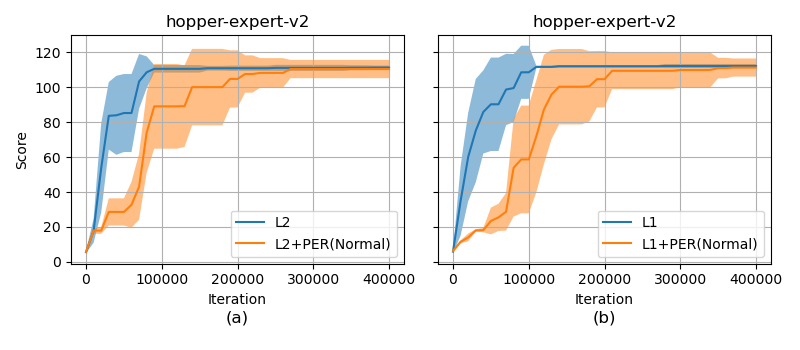}
 \caption{The scores across iterations for policy optimization tested on hopper-expert dataset, where the shaded regions represent the score's mean $\pm$ one standard deviation, the black dotted line indicates the score of IQL (paper).}\label{fig8}
 \end{figure}

  \begin{figure}[H]%
 \centering
 \includegraphics[width=0.8\textwidth]{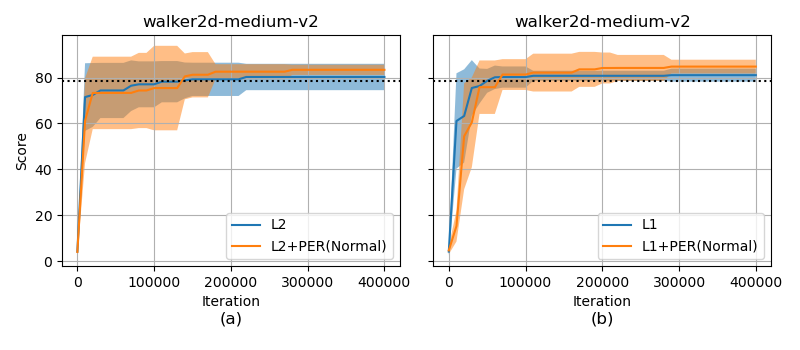}
 \caption{The scores across iterations for policy optimization tested on walker2d-median dataset, where the shaded regions represent the score's mean $\pm$ one standard deviation, the black dotted line indicates the score of IQL (paper).}\label{fig9}
 \end{figure}
 \begin{figure}[H]%
 \centering
 \includegraphics[width=0.8\textwidth]{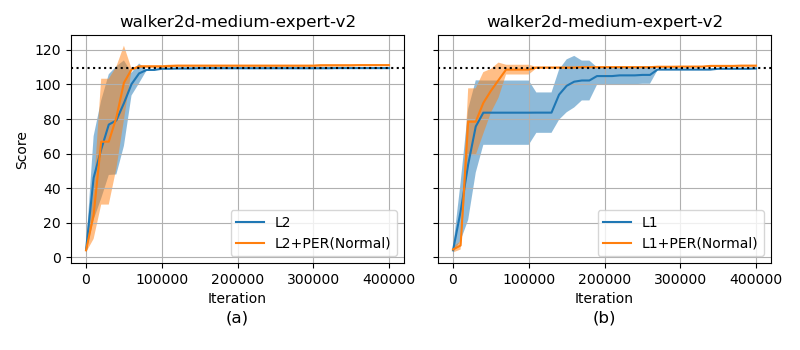}
 \caption{The scores across iterations for policy optimization tested on walker2d-medium-expert dataset, where the shaded regions represent the score's mean $\pm$ one standard deviation, the black dotted line indicates the score of IQL (paper).}\label{fig10}
 \end{figure}
 \begin{figure}[H]%
 \centering
 \includegraphics[width=0.8\textwidth]{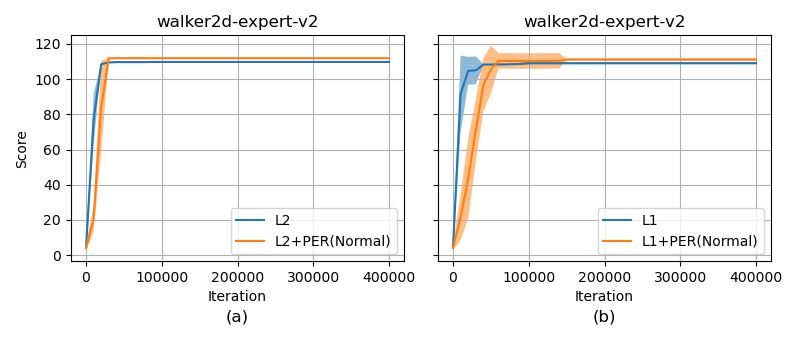}
 \caption{The scores across iterations for policy optimization tested on walker2d-expert dataset, where the shaded regions represent the score's mean $\pm$ one standard deviation, the black dotted line indicates the score of IQL (paper).}\label{fig11}
 \end{figure}

 \begin{figure}[H]%
 \centering
 \includegraphics[width=0.8\textwidth]{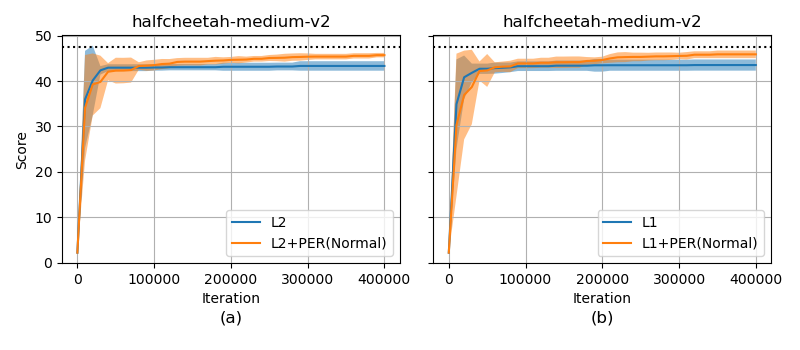}
 \caption{The scores across iterations for policy optimization tested on halfcheetah-medium-expert dataset, where the shaded regions represent the score's mean $\pm$ one standard deviation, the black dotted line indicates the score of IQL (paper).}\label{fig12}
 \end{figure}
 \begin{figure}[H]%
 \centering
 \includegraphics[width=0.8\textwidth]{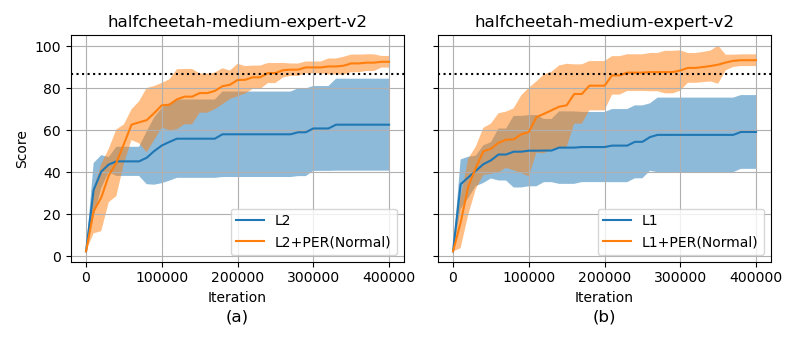}
 \caption{The scores across iterations for policy optimization tested on halfcheetah-medium dataset, where the shaded regions represent the score's mean $\pm$ one standard deviation, the black dotted line indicates the score of IQL (paper).}\label{fig13}
 \end{figure}
 \begin{figure}[H]%
 \centering
 \includegraphics[width=0.8\textwidth]{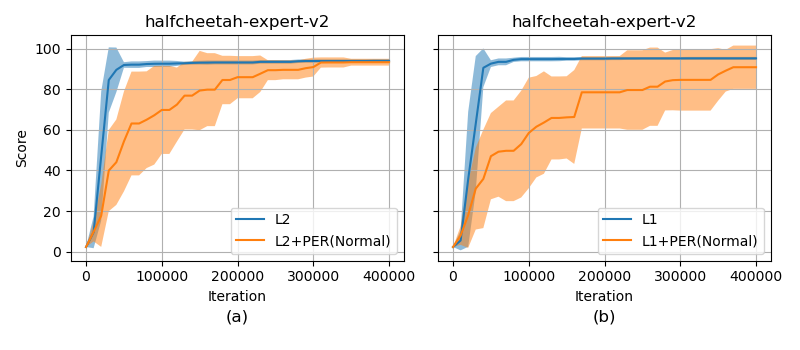}
 \caption{The scores across iterations for policy optimization tested on halfcheetah-expert dataset, where the shaded regions represent the score's mean $\pm$ one standard deviation, the black dotted line indicates the score of IQL (paper).}\label{fig14}
 \end{figure}
 
\end{appendix}

\end{document}